\documentclass[letterpaper, 10 pt, conference]{ieeeconf}  %
\IEEEoverridecommandlockouts
\usepackage{cite}
\usepackage{amsmath,amssymb,amsfonts}
\usepackage{algorithmic}
\usepackage{graphicx}
\usepackage{textcomp}
\usepackage{booktabs}
\usepackage{array}
\usepackage{multirow}
\usepackage{xcolor}
\def\BibTeX{{\rm B\kern-.05em{\sc i\kern-.025em b}\kern-.08em
    T\kern-.1667em\lower.7ex\hbox{E}\kern-.125emX}}

\usepackage{booktabs}
\usepackage[table]{xcolor}

\usepackage{subcaption}
\usepackage[normalem]{ulem} %

\usepackage{amsmath}
\usepackage{hyperref}

\newcommand\gderror[1]{
   \typeout{--------------------------------------------------------------------}
   \typeout{------- #1 ---------}
   \typeout{--------------------------------------------------------------------}
   {\bf #1}
}
\newcounter{gdTmp}
\newcounter{gdLastCount}
\newcommand\maxpage[2][Error]{  %
\ifnum\value{page}>#2
    \gderror{On page {\thepage} we are past page #2 (too long).   #1 }
\else\fi
\setcounter{gdLastCount}{\value{page}} %
}
\newcommand\maxpageSinceLast[2][Error]{  %
\ifnum \numexpr \value{page} - \value{gdLastCount}\relax>#2
    \gderror{Exceeds max length #2 pages. Page \thepage: #1}
\thepage\else\fi
\setcounter{gdLastCount}{\value{page}} %
}

\definecolor{tableorange}{HTML}{FFC6A3}
\definecolor{bestgray}{HTML}{D9D9D9}
\definecolor{secondgray}{HTML}{ECECEC}
\let\labelindent\relax
\usepackage{enumitem}

\usepackage{float}

\usepackage[ruled,vlined,linesnumbered]{algorithm2e}

\begin{document}

\title{Grounding Object-to-Object Hazard Reasoning into 3D Risk Fields for Task-Conditioned Safety}
\author{
Mariana Sosa Guzmán\textsuperscript{1,*},
Yide Song\textsuperscript{2},
Amin Abyaneh\textsuperscript{2},
Gregory Dudek\textsuperscript{1},
Hsiu-Chin Lin\textsuperscript{1,2}
\thanks{$^1$Department of Computer Science, McGill University, Montreal, Canada.}
\thanks{$^2$Department of Electrical and Computer Engineering, McGill University, Montreal, Canada.}
\thanks{*Corresponding author: Mariana Sosa Guzmán
(\texttt{mariana.sosaguzman@mail.mcgill.ca}).}
}

\maketitle

\begin{abstract}
Safety in robotics does not solely depend on geometric constraints, but also on semantics: the given task, environment properties, and interactions between objects in the scene. We present a framework based on vision-language models (VLMs) to infer task- and scene-dependent safety constraints for robot manipulation. We provide the VLM a structured safety tool based on an RGB-D observation and a natural-language task instruction that first decides whether the task should be rejected since its execution may result in a non-recoverable hazardous outcome. Otherwise, it identifies preventable object-pair hazards and specifies the spatial relations that must be avoided. Open-vocabulary segmentation and depth information are then used to ground the detected objects in 3D and construct continuous risk fields that represent the corresponding hazardous regions. These fields are integrated with an obstacle avoidance module that corrects nominal robot motions during execution while accounting for the risk field. We evaluate the framework on preventable hazards, tasks requiring rejection, and safe variants in simulation and the real-world. Project website: \url{https://vlm-hazard-riskfield.github.io}

\end{abstract}

\section{Introduction}

\noindent

Robots that manipulate everyday objects must anticipate if and how their actions can cause harm. Doing so requires connecting what they see with what they are asked to do: an object's properties, its surroundings, and its intended use together determine which interactions are hazardous. Collision avoidance provides an essential foundation \cite{obstacle-avoidance, pot-fields, 1676196, kavraki1996probabilistic}, but geometry alone is far from sufficient for identifying safety constraints. We study the broader challenge of \textit{contextual safety} \cite{brunke-semantic, santos2025updating, ravichandran2026contextual} in robotic manipulation, with the aim of enabling robots to derive safety constraints from the task and scene and utilize visual context and structured reasoning to guide task rejection and motion correction.

\begin{figure}[!t]
    \centering
    \includegraphics[width=1\linewidth]{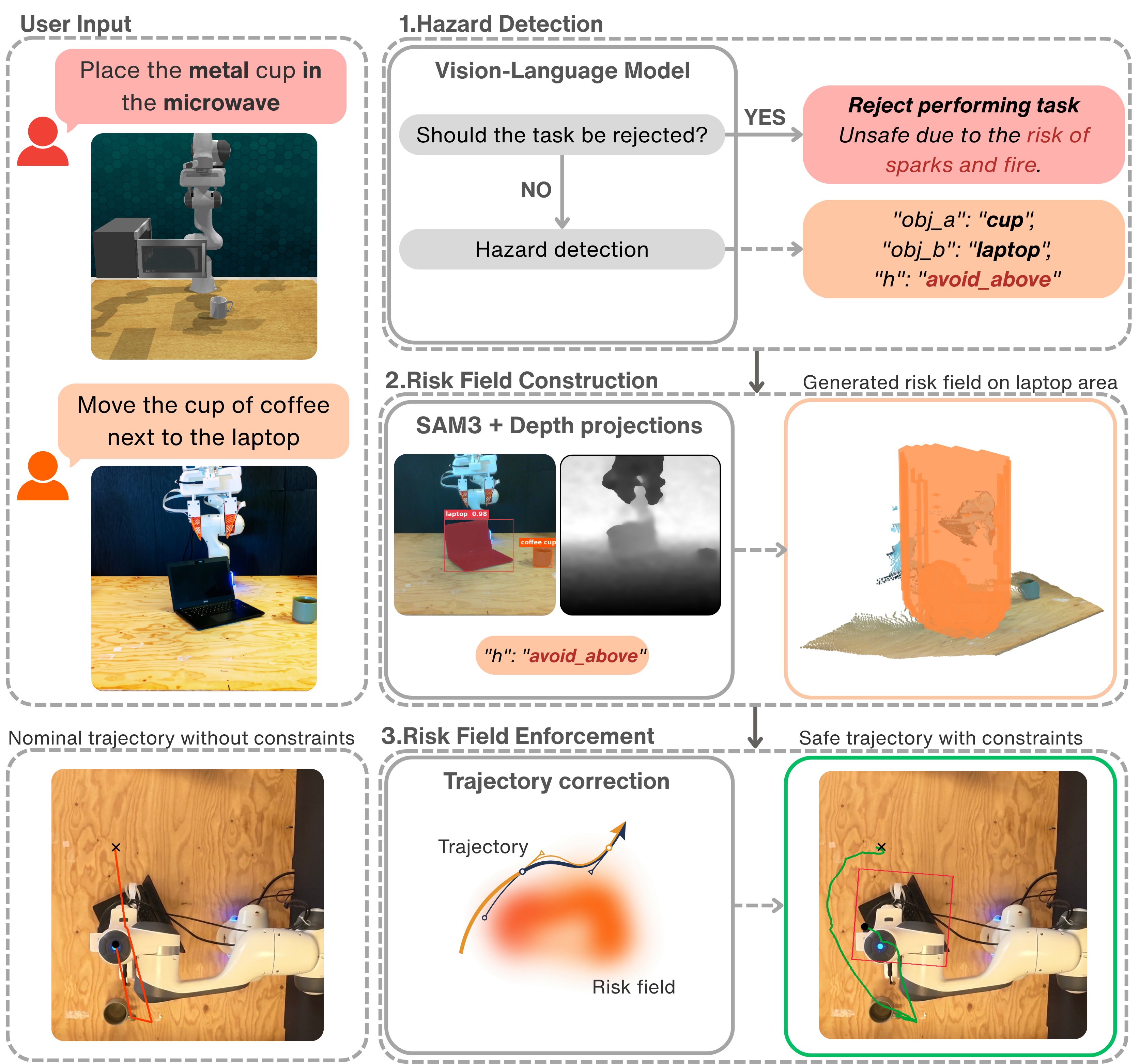}
    \caption{Overview of the proposed framework. Given a task instruction and a scene image, we either reject the task or identifies object-pair hazards (top) that are grounded into a 3D risk field (middle), which guides safe execution (bottom).}
    \label{fig:framework}
\end{figure}

These ideas have been explored both in manipulation \cite{brunke-semantic} and in navigation \cite{santos2025updating, feng2025words, ravichandran2026contextual}. In navigation, constraints are defined over the regions the robot itself may occupy. In manipulation, the robot additionally acts on the environment through the object it holds, so safety depends on the relationship between that object and the remaining objects of the scene. For example, placing a metal cup in a microwave for heating can create a hazard regardless of the transport path, whereas carrying coffee over a laptop creates a hazard that can be avoided by routing the cup around it (Figure.~\ref{fig:framework}). Hazard reasoning must therefore decide whether a task should be performed at all and, if so, which regions become hazardous for the manipulated object due to the task.

In prior work, this knowledge typically comes from a human who specifies the constraints \cite{santos2025updating, feng2025words, plug-safe-chip, amani2025bayesian}
or from Hamilton-Jacobi reachability \cite{reach-analysis}. However, a predefined hazard set could not cover all possible hazards that arise in an open-world environment. There is increasing interest in extracting this contextual knowledge directly from large language models (LLMs). Existing work in manipulation segments visual input into a semantic map, uses a pretrained model to reason about safe and unsafe behavior, and enforces the resulting constraints through control barrier functions (CBFs) \cite{brunke-semantic}; however, reducing the scene to a language representation limits the context available for reasoning. Vision-language models (VLMs) jointly process semantic and visual information, allowing reasoning over objects, attributes, and relationships within a scene. This direction has been explored for navigation \cite{ravichandran2026contextual}, but its application to hazards induced by object interactions in manipulation remains underexplored.

Knowing what is unsafe, however, does not by directly prevent it. Safety filters such as CBFs \cite{cbf, cbf2} provide formal guarantees that corrected nominal trajectories to keep the robot safe with respect to spatially unsafe sets, under their respective modeling and control assumptions. These sets and rules must be specified in advance. %
Foundation models inherit a high-level semantic understanding to reason about potential hazard interactions, but their predictions need grounding to be applied.
A hazard detection pipeline must therefore bridge these two ideas: it must reason about safety based on the environment and task, and ground the predicted hazards in a representation that a classical filter can act on continuously.

In this work, we propose a framework centered around a VLM with visual context  and a text description of the task to detect hazards arising from interactions between objects and induced by the robot's actions. A structured schema-based toolset first determines whether the task should be rejected and, otherwise, outputs object-pair hazards with geometric restrictions. Open-vocabulary segmentation and depth ground these restrictions in an automatically generated 3D risk field, which an obstacle modulation filter uses to correct the manipulated object's motion. In parallel, we present a systematic study of how context representation and structured reasoning affect pretrained models for object-to-object hazard detection.
Our contributions are:
\begin{itemize}
    \item A structured VLM safety toolset that separates task rejection from identifying preventable object-pair hazards, and outputs geometric restrictions that can be grounded in the scene.

    \item An automatically generated 3D risk field, built from open-vocabulary segmentation and depth, that represents non-convex hazard geometry without fitting shape primitives, and connects VLM hazard reasoning to obstacle modulation.

    \item A study of context representation and structured hazard reasoning, identifying false-rejection failures and evaluating downstream safe execution in simulation and real-world manipulation.

\end{itemize}

\noindent
We evaluate our methodology in simulation and real-world experiments, comparing it against task-only reasoning, reasoning over scene descriptions, and geometric obstacle modulation. The evaluation covers tasks requiring rejection, preventable hazards, and safe variants, assessing both hazard avoidance and task completion.

\section{Related Work}

\subsection{Motion Safety Filters}
\noindent
\label{sec:related-safety-filter}
In classical motion planning, safety has often been synonymous of \textit{collision and obstacle avoidance}, usually handled by reactive methods like potential fields \cite{pot-fields,coulombe.icra.2023, 9765824}, which mainly focus on maintaining a distance between the robot and obstacles while it moves towards a goal. This concept builds on the idea of enforcing safety by modifying the robot's motion whenever the nominal plan may lead to an unsafe state, motivating the development of \textit{safety filters}, which act as modules between a nominal planner and the robot's low-level controller, correcting behavior to prevent violations of predefined safety constraints. Methods such as Control Barrier Functions (CBFs) \cite{cbf, cbf2}, Model Predictive Control (MPC) \cite{mpc, mpc2}, and Hamilton-Jacobi Reachability Analysis \cite{reach-analysis} provide formal guarantees to restrict unsafe behavior regarding a predefined safe/unsafe set, while other methods modulate robot trajectories with stability~\cite{abyaneh2024globally, emp} or contraction~\cite{partialcont, abyaneh2025contractive} guarantees for safer execution.

\subsection{Foundation models for safety}

\noindent
The use of foundation models for safety has grown in different directions. Affordance-based methods use LLMs to decompose tasks into subtasks and select actions based on contextual feasibility \cite{saycan, voxposer, llm+a}, reasoning about how objects and the environment constrain valid interactions, though these approaches struggle with partially observable scenes. Rule-based methods instead rely on hard-coded constraints, which improves interpretability but limits scalability \cite{10801576, safeplan, selp}. Other work targets uncertainty estimation, using language-based confidence to flag unreliable predictions \cite{sattar2014reducing, 6907842, kadavath2022language}, sometimes triggering clarification requests at the cost of reduced autonomy \cite{knowno}. More recently, LLM-as-judge strategies use a model to evaluate and critique a proposed plan \cite{10632622, semantic-anomaly-2023, gu2024survey}, with some approaches splitting the roles across two models, one as planner, one as judge \cite{11246041}.

\subsection{Contextual Safety}
\noindent
Recently, some approaches have focused on using foundation models for solving more nuanced safety challenges. ``Semantic safety" is introduced in \cite{brunke-semantic}, where an LLM reasons over scene labels obtained from SAM and CLIP to infer object-level hazard constraints, which are then converted into superquadric-based keep-out regions and enforced with a CBF-QP filter; since the LLM reasons only over the extracted labels, the context available to it is bounded by the label set. \cite{ravichandran2026contextual} (CORE) instead derive hazard regions directly from the live image, but their formulation of hazard is environment-derived and do not condition on the sub-sequential interaction of elements in the environment or the task being performed by the robot, so the same scene yields the same hazard field regardless of what the robot is doing.

Our work focuses on hazard reasoning that combines semantic knowledge with visual grounding, conditioning on the specific task and environment. We use a VLM to reason over the scene image and task description to first determine if the task has to be rejected based on non-recoverable outcomes it may lead; if the task is not rejected, the reasoning model propose candidate hazards, used to build a risk field that interfaces with a classical controller.

\section{Problem Statement}
\label{problem_statement}

\newcommand{\vnom}{\mathbf{v}_{\mathrm{nom}}}
\newcommand{\task}{T}                         %
\newcommand{\obs}{I}                          %
\newcommand{\Tcw}{\mathbf{T}_{\mathrm{world\_cam}}}
\newcommand{\vlm}{\Psi}     %
\newcommand{\rej}{\delta}                   %
\newcommand{\hazset}{\mathcal{H}}           %
\newcommand{\obja}{a}           %
\newcommand{\objb}{b}           %
\newcommand{\spatrel}{\mathcal{R}}
\newcommand{\field}{\rho}                 %
\newcommand{\vmod}{\mathbf{v}_{\mathrm{mod}}}

\noindent
Given a natural-language task instruction $\task$, a robot manipulator operating in a scene, whose end-effector position $\mathbf{x} \in \mathbb{R}^3$ evolves as $\dot{\mathbf{x}} = \mathbf{v}$, a nominal controller producing a velocity command $\vnom = f(\mathbf{x})$, single RGB-D observation $\obs$ of the scene taken at the initial state, and a pre-trained VLM model $\vlm$. We assume the nominal controller has no notion of hazards. The objects in the scene are not known a priori, and no predefined safe or unsafe sets are given.

With the above information, our goal is to (1) determining whether $\task$ should be executed at all and (2) if $\task$ can be performed with appropriate precautions, obtain a hazard-free trajectory that completes the task.

\section{Methodology}

\newcommand{\fieldfigure}{
\begin{figure}[t]
    \centering
    \captionsetup[subfigure]{labelformat=empty}
    \begin{subfigure}[t]{0.49\linewidth}
        \centering

        \includegraphics*[width=\linewidth]{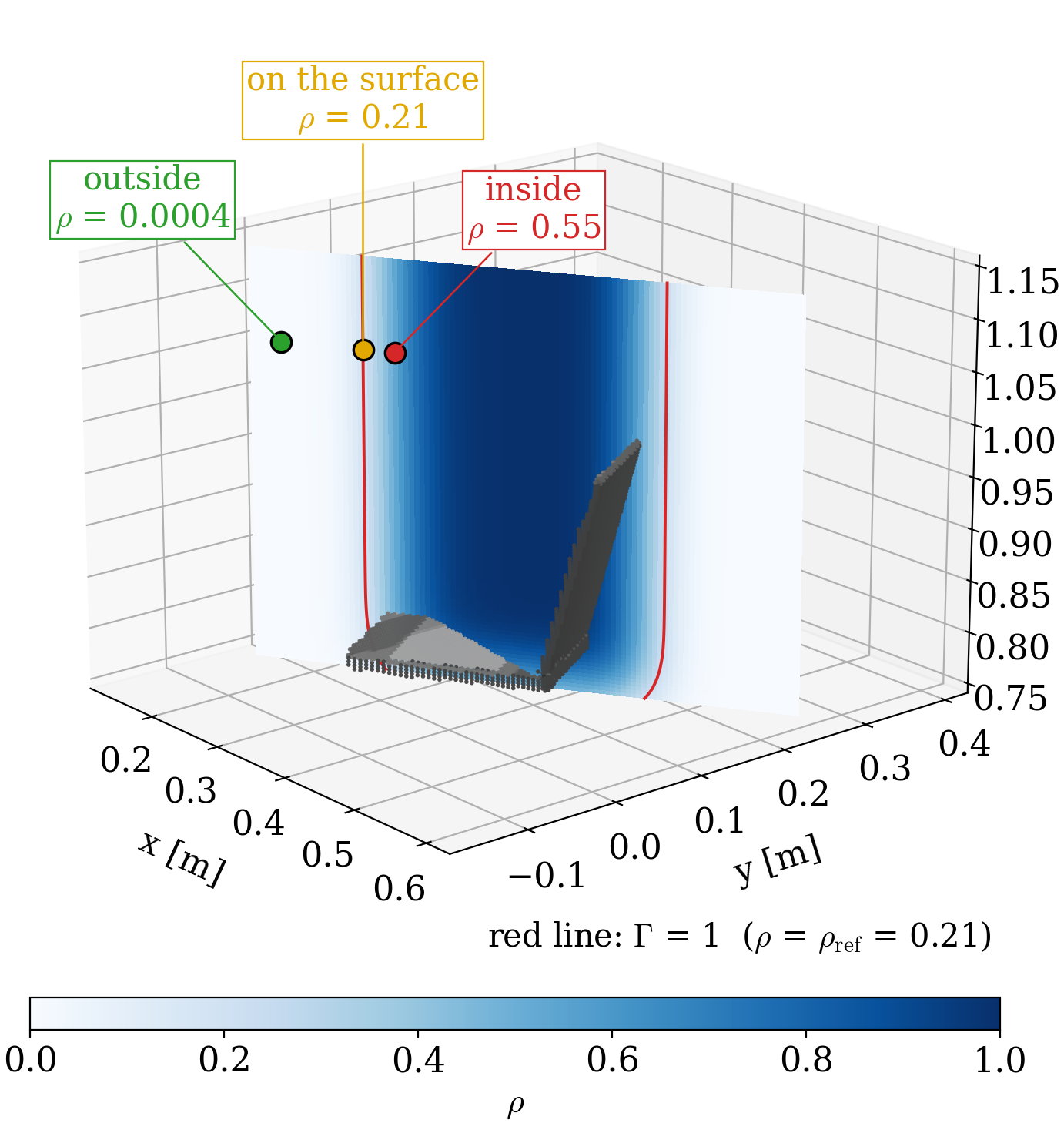}

        \caption{Density $\rho$ for the laptop.}
        \label{fig:risk_field_rho}
    \end{subfigure}
    \hfill
    \begin{subfigure}[t]{0.49\linewidth}
        \centering

        \includegraphics*[width=\linewidth]{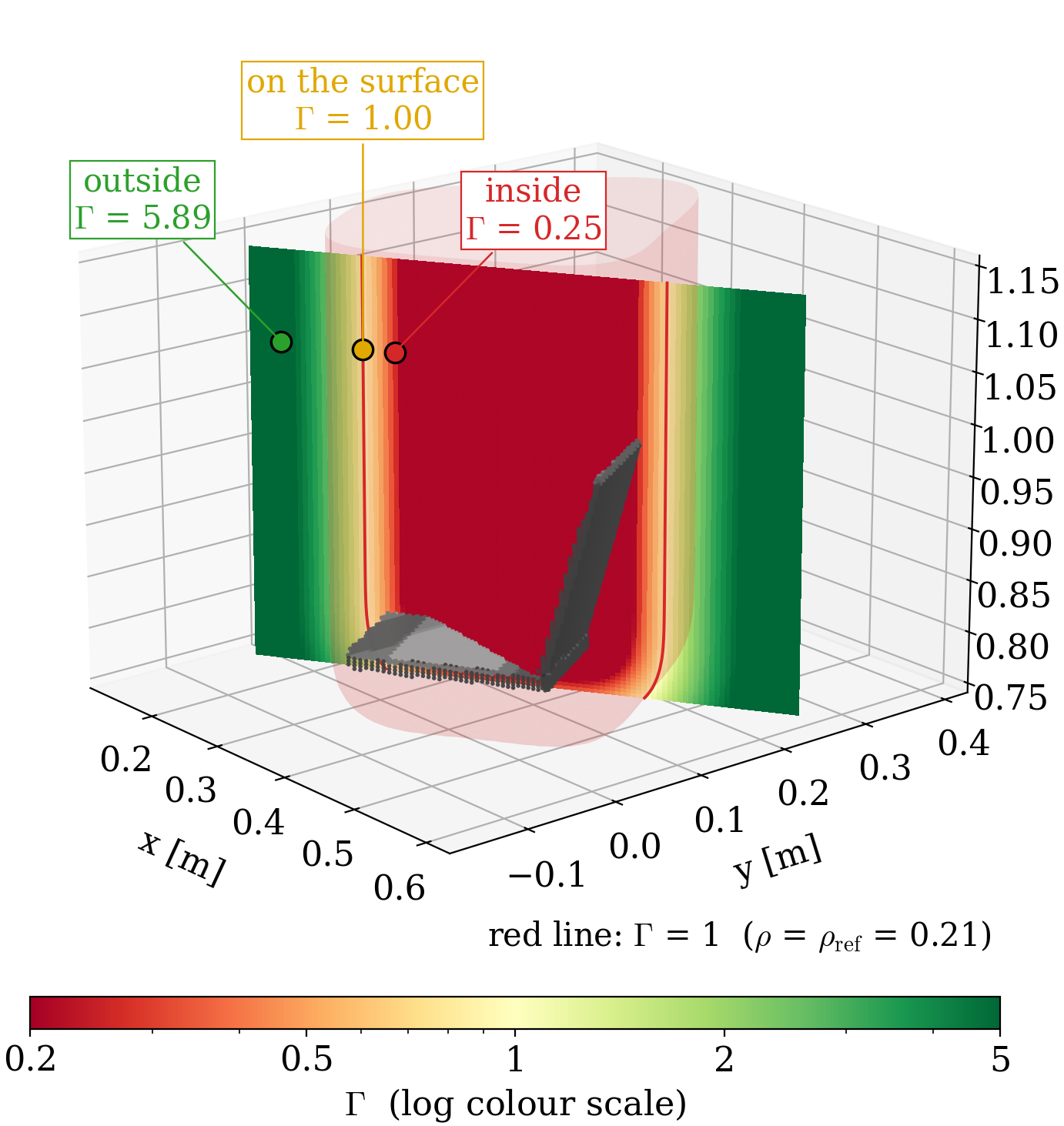}
        \caption{$\Gamma$ for the laptop.}
        \label{fig:risk_field_gamma}
    \end{subfigure}
    \caption{Density and gamma for laptop}
    \label{fig:risk_field}
\end{figure}
}
\noindent
In this section, we present the hazard detection framework using visual context and task description.

\subsection{Hazard Reasoning}
\label{sec:hazard-reasoning}

\noindent

\noindent
To solve the questions stated in Section~\ref{problem_statement}, we divide the hazard reasoning problem into two problem categories: non-recoverable hazards and recoverable hazards, as shown in Figure~\ref{fig:framework}.

\textbf{I. Non-recoverable hazards.} Some user inputs are inherently unsafe regardless of how the trajectory is planned because the hazard arises from the task itself. For example, \textit{"placing a metal cup in a microwave"} (Fig.~\ref{fig:framework} top) is a fire hazard and should never be executed. Therefore, we first need to evaluate
\begin{equation}
  \rej = \vlm(\obs, \task) \in \{0, 1\},
\label{eq:rejection_check}
\end{equation}
where $\vlm$ judges whether executing $\task$ in the observed scene leads to
a non-recoverable unsafe state, and returns the decision, where $\rej = 1$ means a rejected task.

\textbf{II. Recoverable hazards}. For tasks that are not refused, $\vlm$ determines if there are any preventable hazards present (e.g., a spilling hazard can be prevented by not moving liquid containers above electronic devices), and infers them from $\task$ and the observation $\obs$.
Each hazard in the set of inferred hazards $\hazset$ is represented as a tuple:
\begin{equation}
  h_i = (\obja_i, \objb_i, c_i, r_i) \in \hazset
  \label{equ:inferred-hazard}
\end{equation}
where $a_i$ and $b_i$ are the two objects whose interaction produces the hazard, with $\objb$ being the anchor hazardous object, $c_i$ is a hazard description, and $r_i$ is the spatial relation to avoid between them, which can be: \textit{``avoid\_above'', ``avoid\_below'', ``avoid\_inside'', ``avoid\_between'', ``avoid\_collision''}.

For example, the user input ``move the cup of coffee next to the laptop", shown in Figure \ref{fig:framework}, should be done carefully to avoid spilling liquid on the laptop; the equation in \ref{equ:inferred-hazard} becomes
$h_i = (\textit{"cup"}, ~\textit{"laptop"}, ~\textit{"spill\_hazard"}, ~\textit{"avoid\_above"})$.

We formulate the reasoning process as schema-based function calling.
Specifically, we provide the pre-trained model with a set of specific schema-based function definitions, each characterized by a name, a description, and arguments. Given the task and scene observation, the model selects the appropriate function and generates its corresponding arguments. We define one function for non-recoverable hazards and a set of functions for recoverable hazards. Unlike free-form prompting, which specifies to the model what the output should look like, our tool set defines which safety functions it can select and what arguments it should provide. Previous work demonstrates that language models could select user-defined tools \cite{pmlr-v267-patil25a}. In our case, the VLM uses native function calling to express safety decisions by calling a predefined tool in our toolset.

In the example above, instead of expressing \textit{"avoid moving above the laptop"} in natural language, the VLM took the \textit{"avoid\_above"} function, then passed the cup as function argument $\obja$ (the manipulated object) and the laptop as $\objb$ (the anchor hazard object).

\subsection{Grounding and Risk Field Construction}
\label{subsec:detection}

\noindent
To ground semantic hazard information and enforce it through a safety filter, we focus on every pair of $\obja_i$ and $\objb_i$ related to a detected hazard $\hazset$ in the scene to construct a risk field.

First, for each pair of objects $\obja_i, \objb_i$, we obtain each object's segmentation mask by using SAM-3 \cite{carion2026sam}. These are combined with the registered depth data captured at the initial scene state, and the masked pixels are back-projected into the world frame via the standard pinhole model using the camera's intrisic matrix and frame transformation with the extrinsics matrix \cite{slam-cam, opencv}.

The set of world-frame points obtained for each detected object builds its point cloud $\mathcal{P}$. Using the segmentation mask with the geometric restriction $r$ from the hazard schema, we transform $\objb$'s point cloud into a hazard region. For example, to include the \textit{``avoid\_above"} restriction, the point cloud is extruded vertically from the surface to a fixed maximum height, forming a pillar. %
For \textit{``avoid\_collision"}, no extrusion is added, and the object itself is considered the risk region.

Since our downstream safety filter requires a continuous field rather than a discrete point cloud, we accumulate the raw hazard points on a space of 3D voxels, where each voxel has a count of hazard points. We apply Gaussian smoothing to the voxelized space to generate the field $\field({\mathbf{p}})$ for any point $ \mathbf{p} \in \mathcal{P}$.

\newcommand{\gamme}{collision distance function}

For this obstacle avoidance (Sec.~\ref{sec:related-safety-filter}), a continuous function $\Gamma (\mathbf{p})$ is required to describe a point $\mathbf{p}$ with respect to the hazard surface and satisfies:
\begin{equation}
    \Gamma(\mathbf{p})
    \begin{cases}
        > 1 &\text{Outside of obstacles}\\
        = 1 &\text{On the surface of obstacles}\\
        <1 &\text{Inside of obstacles}\\
    \end{cases}
\end{equation}
Therefore, we need to convert $\field$ into $\Gamma (\mathbf{p})$.  A density value $\field(\mathbf{p})$ alone cannot tell whether a point $\mathbf{p}$ lies on, inside, or outside the hazard surface, so we need a reference surface density $\field_{ref}$, defined in Equation \ref{eq:o_ref}.

\begin{equation}
    \field_{ref} = \beta \cdot \mathrm{median}_{\mathbf{p} \in \mathcal{P} } \, \field(\mathbf{p}),
\label{eq:o_ref}
\end{equation}
where $\beta\in(0,1]$ scales the surface occupancy level of raw hazard points so that $\Gamma$ = 1 on the hazard surface. We use the median instead of the mean since the Gaussian smoothing process dismisses points near the edges of the point cloud, where fewer points contribute to the local density \cite{jones1993simple}. Therefore, we define $\Gamma$ for any point $\mathbf{x}$ as

\begin{equation}
\Gamma(\mathbf{x}) = 1 + \frac{\ln(\field_{ref}/\field(\mathbf{x}))}{\kappa}.
\label{eq:gamma}
\end{equation}

\begin{figure}[t]
    \centering
    \includegraphics[width=0.9\linewidth]{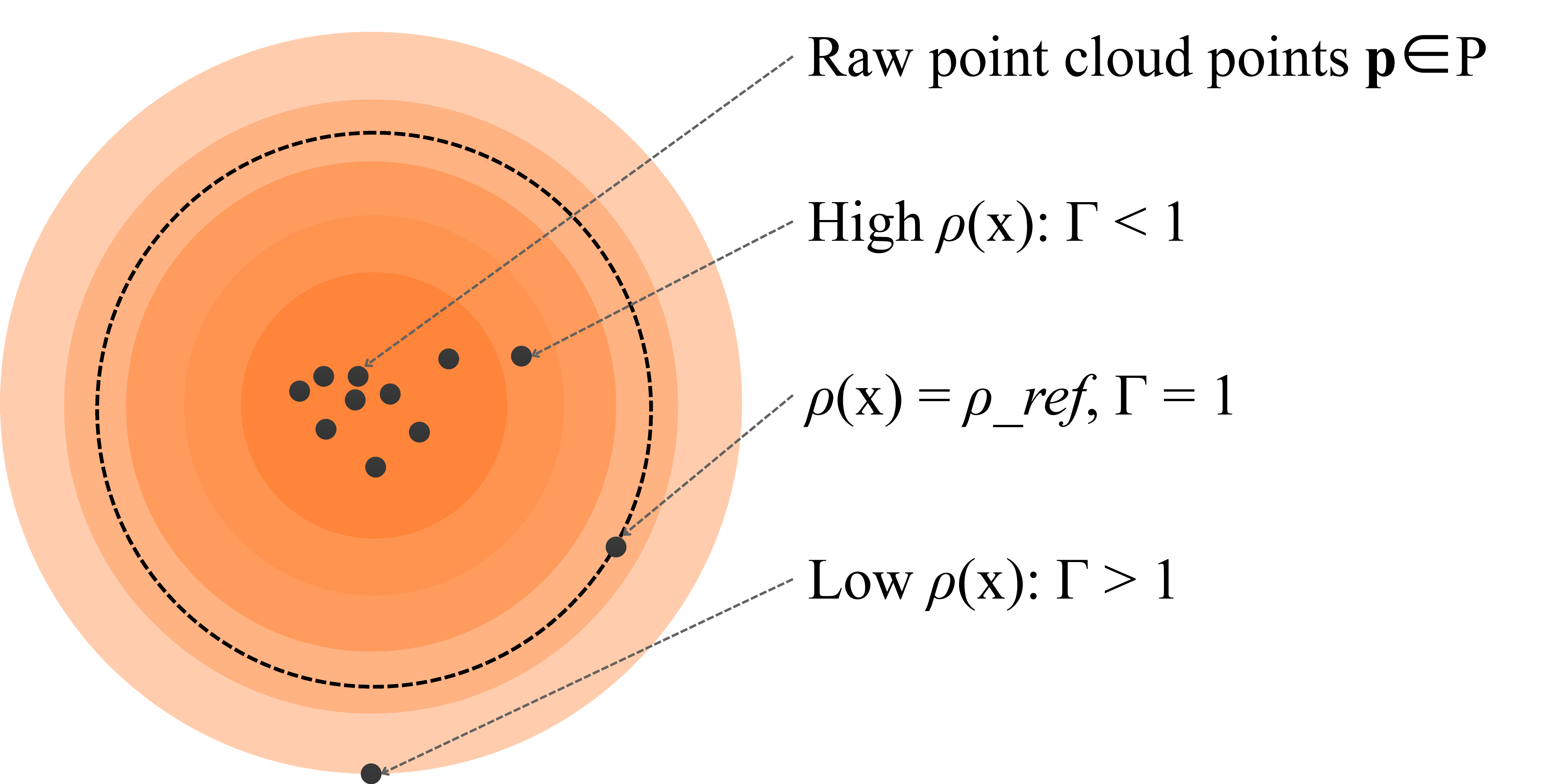}
    \caption{Representation of how the location of a point $x$ in the field $\field$ affects $\Gamma$.
    }
    \label{fig:placeholder}
\end{figure}

\noindent
Where $\kappa$ is a hyperparameter to control the influence range of the field. Now, wherever $\field(\mathbf{x}) = \field_{ref}$, the ratio is 1, therefore $\ln(\field_{ref}/\field(\mathbf{x}))=0$, satisfying the boundary condition the modulation filter requires.
An example is illustrated in Fig.~\ref{fig:placeholder}. With this modification, $\Gamma<1$ in regions of high $\field(\mathbf{p})$ and $\Gamma>1$ is regions of low $\field(\mathbf{p})$.

\fieldfigure

\subsection{Risk Field Integration with Obstacle Modulation}

\noindent
As described in the previous section, we use obstacle modulation~\cite{coulombe.icra.2023} to enforce the constraints defined in the hazard schema, where $\obja$ is the manipulated object associated with the hazard, and $\objb$ denotes the base of the risk field.

Evaluating $\Gamma$ at the object position is insufficient, since $\obja$ has a shape and volume; the end-effector point position may remain outside the hazard region while another part of the object does not.
To account for this, we adapt the formulation of ~\cite{coulombe.icra.2023}, which was designed for collision avoidance between two convex objects, and we are adapting it for collision avoidance between a bounding box and a risk field.

\begin{figure}[t]
    \centering
    \includegraphics[width=0.8\linewidth]{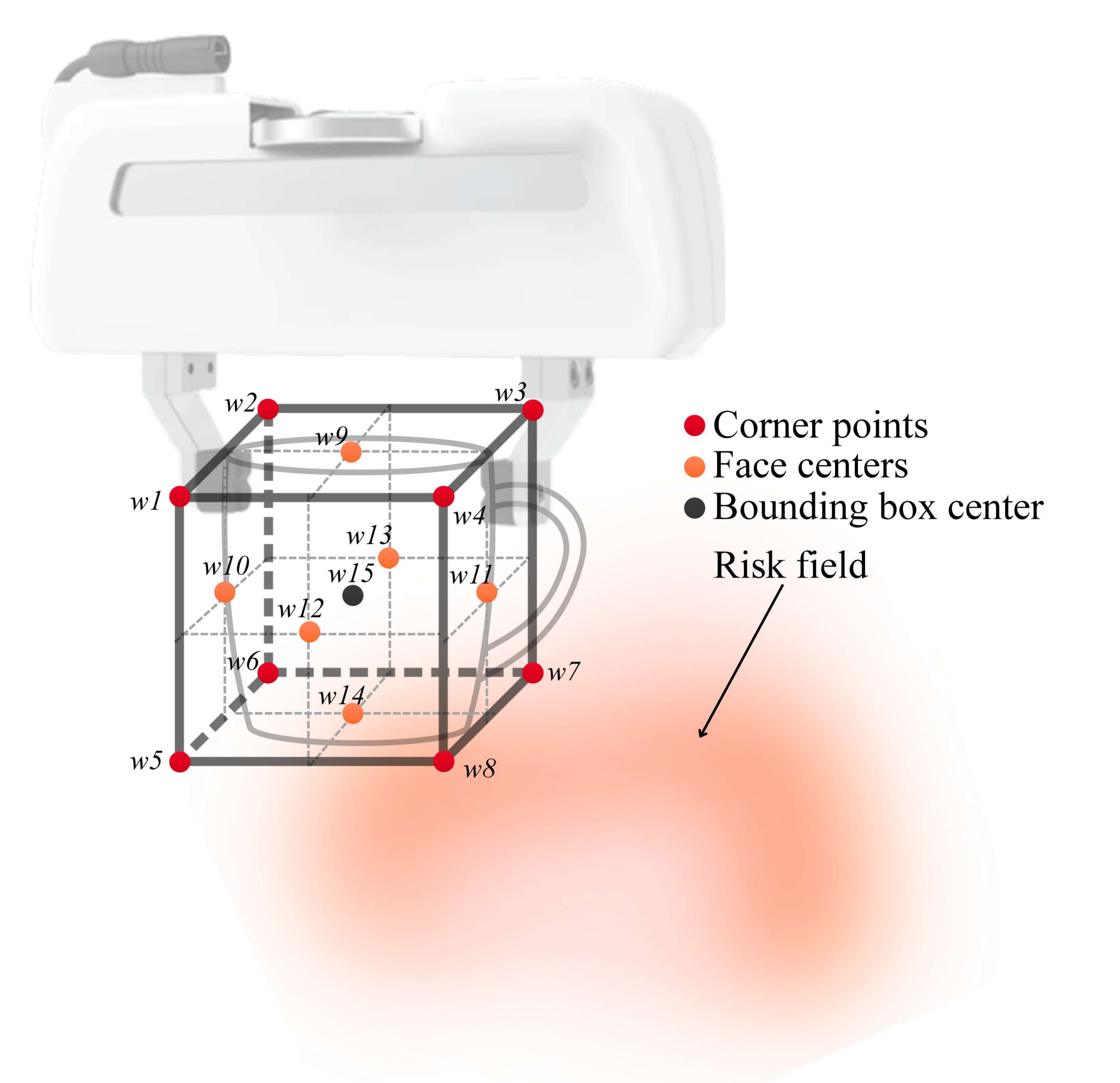}
    \caption{Example of $\Gamma$ estimation between a bounding box of a mug $\obja$ and a risk field $\field$.
    The corner points, face centers, center of the bounding-box are evaluated to determine whether any part of the object enters the hazardous region.}
    \label{fig:aabb-points}
\end{figure}

Specifically, the segmentation mask and the corresponding 3D points of $\obja$ obtained as in Section~\ref{subsec:detection} are used to build an axis-aligned bounding box of the object,
characterized by the corner points and face centers ($\mathbf{w}_1, \mathbf{w}_2, \dots, $). We use this set of points to evaluate $\Gamma$, as shown in Figure \ref{fig:aabb-points}, and take the minimum $\Gamma$ value across these points as the object's proximity to the hazard region.
\begin{equation}
    \Gamma_{a,b} = \min \Gamma (\mathbf{w}_i)
\end{equation}
\noindent This $\Gamma_{a,b}$ will be used to compute the obstacle modulation function in ~\cite{coulombe.icra.2023}.

\section{Experiments}
\label{exp_section}

\newcommand{\stat}[2]{%
  \shortstack[c]{#1\\[-1.5pt]{\color{gray}\tiny$\pm$#2}}%
}

\newcommand{\executiontable}{%
\begin{table*}[t]
\centering
\caption{Evaluation of safe trajectory execution. Dark gray shading indicates the best result in each column. Top arrows indicate whether higher or lower values are preferred, respectively.}
\label{tab:execution}
\begin{minipage}{\linewidth}
\centering

\end{minipage}
\par\vspace{3pt}

\setlength{\tabcolsep}{1.0pt}
\setlength{\extrarowheight}{1.5pt}

\begin{tabular*}{\textwidth}{
@{\hspace{6pt}\extracolsep{\fill}}
l |ccc | ccc | ccc | cc | cc | cc | ccc
@{\hspace{6pt}}
}

\toprule

\rowcolor{tableorange}[6pt][0pt]
& \multicolumn{9}{@{}c@{}}{\textbf{Preventable hazard}}
& \multicolumn{6}{@{}c@{}}{\textbf{Safe variant}}
& \multicolumn{3}{@{}c@{}}{\textbf{Overall}} \\[3pt]

\specialrule{0.3pt}{0pt}{0pt}

\rowcolor{tableorange}[6pt][0pt]
& \multicolumn{3}{@{}c@{}}{\textbf{Drawer-plant}}
& \multicolumn{3}{@{}c@{}}{\textbf{Laptop-Cup}}
& \multicolumn{3}{@{}c@{}}{\textbf{Steak-Dripping}}

& \multicolumn{2}{@{}c@{}}{\textbf{Drawer-Safe}}
& \multicolumn{2}{@{}c@{}}{\textbf{Laptop-Cube}}
& \multicolumn{2}{@{}c@{}}{\textbf{Steak-Safe}}

& \multicolumn{3}{@{}c@{}}{\textbf{Average}} \\[3pt]

\specialrule{0.3pt}{0pt}{0pt}

\rowcolor{tableorange}[6pt][6pt]
\textbf{Method}
& \textbf{SR $\uparrow$}
& \textbf{SSR $\uparrow$}
& \textbf{Viol.\ $\downarrow$}

& \textbf{SR $\uparrow$}
& \textbf{SSR $\uparrow$}
& \textbf{Viol.\ $\downarrow$}

& \textbf{SR $\uparrow$}
& \textbf{SSR $\uparrow$}
& \textbf{Viol.\ $\downarrow$}

& \textbf{SR $\uparrow$}
& \textbf{SSR $\uparrow$}

& \textbf{SR $\uparrow$}
& \textbf{SSR $\uparrow$}

& \textbf{SR $\uparrow$}
& \textbf{SSR $\uparrow$}

& \textbf{SR $\uparrow$}
& \textbf{SSR $\uparrow$}
& \textbf{Viol.\ $\downarrow$} \\[3pt]

\midrule

No context
& \stat{33.3}{15.3}
& \stat{0.0}{0.0}
& \stat{4.1}{1.3}

& \stat{63.3}{5.8}
& \stat{46.7}{5.8}
& \stat{3.8}{2.8}

& \cellcolor{bestgray}\stat{100.0}{0.0}
& \stat{0.0}{0.0}
& \stat{11.4}{0.1}

& \stat{26.7}{5.8}
& \stat{26.7}{5.8}

& \cellcolor{bestgray}\stat{100.0}{0.0}
& \cellcolor{bestgray}\stat{100.0}{0.0}

& \cellcolor{bestgray}\stat{100.0}{0.0}
& \cellcolor{bestgray}\stat{100.0}{0.0}

& \stat{70.6}{4.2}
& \stat{45.6}{1.0}
& \stat{6.5}{1.4}
\\[3pt]

Text only
& \stat{36.7}{11.5}
& \stat{0.0}{0.0}
& \stat{5.4}{1.7}

& \stat{23.3}{11.5}
& \stat{6.7}{5.8}
& \stat{5.0}{1.4}

& \cellcolor{bestgray}\stat{100.0}{0.0}
& \stat{0.0}{0.0}
& \stat{11.5}{0.3}

& \stat{3.3}{5.8}
& \stat{3.3}{5.8}

& \cellcolor{bestgray}\stat{100.0}{0.0}
& \cellcolor{bestgray}\stat{100.0}{0.0}

& \stat{97.0}{5.2}
& \stat{97.0}{5.2}

& \stat{60.1}{3.4}
& \stat{34.5}{1.0}
& \stat{7.3}{1.1}
\\[3pt]

Simple OA
& \stat{100.0}{0.0}
& \stat{0.0}{0.0}
& \stat{27.5}{0.0}

& \stat{40.0}{0.0}
& \stat{10.0}{0.0}
& \stat{6.7}{0.0}

& \cellcolor{bestgray}\stat{100.0}{0.0}
& \stat{0.0}{0.0}
& \stat{11.5}{0.0}

& \cellcolor{bestgray}\stat{100.0}{0.0}
& \cellcolor{bestgray}\stat{100.0}{0.0}

& \cellcolor{bestgray}\stat{100.0}{0.0}
& \cellcolor{bestgray}\stat{100.0}{0.0}

& \cellcolor{bestgray}\stat{100.0}{0.0}
& \cellcolor{bestgray}\stat{100.0}{0.0}

& \stat{90.0}{0.0}
& \stat{51.7}{0.0}
& \stat{15.2}{0.0}
\\[3pt]

\midrule

\textbf{VLM (ours)}
&  \cellcolor{bestgray}\stat{100.0}{0.0}
& \cellcolor{bestgray}\stat{80.0}{0.0}
& \cellcolor{bestgray}\stat{3.1}{1.3}

& \cellcolor{bestgray}\stat{70.0}{0.0}
& \cellcolor{bestgray}\stat{60.0}{0.0}
& \cellcolor{bestgray}\stat{2.5}{0.0}

& \stat{75.8}{5.2}
& \cellcolor{bestgray}\stat{57.6}{5.2}
& \cellcolor{bestgray}\stat{7.5}{1.3}

&\cellcolor{bestgray} \stat{100.0}{0.0}
& \cellcolor{bestgray}\stat{100.0}{0.0}

& \cellcolor{bestgray}\stat{100.0}{0.0}
&\cellcolor{bestgray} \stat{100.0}{0.0}

& \stat{90.9}{0.0}
& \stat{90.9}{0.0}

& \stat{89.4}{0.9}
& \cellcolor{bestgray}\stat{81.4}{0.9}
& \cellcolor{bestgray}\stat{4.4}{0.4}
\\[3pt]

\bottomrule

\end{tabular*}

\end{table*}
}

\newcommand{\rejectiontablenew}{%
\begin{table}[t]
\centering
\caption{Rejection tasks study with different ablations. Dark gray shading indicates the best result in each column. Top arrows indicate whether higher or lower values are preferred, respectively.}

\label{tab:ablation_rej}
\setlength{\tabcolsep}{2.5pt}
\setlength{\extrarowheight}{2pt}
\begin{tabular}{lccc}
\toprule
\rowcolor{tableorange}
& \multicolumn{1}{c}{\textbf{\shortstack{Correct\\rejection}} $\uparrow$}
& \multicolumn{2}{c}{\textbf{\shortstack{False\\rejection}} $\downarrow$}
\\
\noalign{%
  {\color{tableorange}\hrule height 2.0pt}%
  {\color{black}\hrule height 0.4pt}%
  {\color{tableorange}\hrule height 2.0pt}%
}
\rowcolor{tableorange}
\textbf{Method}
& \textbf{\shortstack{Non-recoverable\\hazard}}
& \textbf{\shortstack{Recoverable\\hazard}}
& \textbf{\shortstack{Safe\\variants}}
\\
\midrule
VLM + Task
& \cellcolor{bestgray}96/96
& 63/93
& 42/90
\\
\shortstack[l]{VLM + Task + Scene Text}
& \cellcolor{bestgray}96/96
& 83/93
& 58/90
\\
 VLM + Task + Scene Image
& \cellcolor{bestgray}96/96
& 78/93
& 49/90
\\
\shortstack[l]{VLM + Task + Scene Text \\ + Tool Calling }
& \cellcolor{bestgray}96/96
& 24/93
& 30/90
\\
\midrule
\textbf{\shortstack[l]{VLM + Task + Scene Image\\+ Tool Calling (Ours)}}
& \cellcolor{bestgray}96/96
& \cellcolor{bestgray}0/93
& \cellcolor{bestgray}0/90
\\
\bottomrule
\end{tabular}

\end{table}
}%

\newcommand{\rejectiontable}{%
\begin{table}[t]
\centering
\caption{\textbf{Rejection ablation.}}
\label{tab:ablation}

\setlength{\tabcolsep}{4pt}
\setlength{\extrarowheight}{2pt}
\footnotesize

\begin{tabular}{lccccc}

\toprule

\rowcolor{tableorange}
&
\multicolumn{3}{c}{\textbf{Correct rejection} $\uparrow$}
&
\multicolumn{2}{c}{\textbf{False rejection} $\downarrow$}
\\

\specialrule{0.3pt}{0pt}{0pt}

\rowcolor{tableorange}
\textbf{Method}
& \textbf{Fork}
& \textbf{Soap}
& \textbf{Micro.}
& \textbf{\shortstack{Safe\\scenes}}
& \textbf{\shortstack{Preventable\\hazard}}
\\

\midrule

No context
& \cellcolor{bestgray}33/33
& \cellcolor{bestgray}33/33
& \cellcolor{bestgray}30/30
& 42/90
& 41/93
\\

Text only
& \cellcolor{bestgray}33/33
& \cellcolor{bestgray}33/33
& \cellcolor{bestgray}30/30
& 54/90
& 78/93
\\

VLM caption
& \cellcolor{bestgray}33/33
& \cellcolor{bestgray}33/33
& \cellcolor{bestgray}30/30
& 1/90
& \cellcolor{bestgray}0/93
\\

\midrule

\textbf{VLM (ours)}
& \cellcolor{bestgray}33/33
& \cellcolor{bestgray}33/33
& \cellcolor{bestgray}30/30
& \cellcolor{bestgray}0/90
& \cellcolor{bestgray}0/93
\\

\bottomrule

\end{tabular}
\end{table}
}

\newcommand{\backbonesweeptable}{

\begin{table}[t]
\centering
\caption{Our Toolset vs.\ state-of-the-art VLM. Dark and light gray indicate the best and second-best in each column, respectively. Top arrows indicate whether higher or lower values are preferred, respectively.}
\label{tab:backbone_sweep}

\setlength{\tabcolsep}{2pt}
\setlength{\extrarowheight}{1pt}
\footnotesize

\begin{tabular}{l c ccc ccc}

\toprule

\rowcolor{tableorange}
\textbf{VLM}
& \textbf{Cost }
& \multicolumn{3}{c}{\textbf{Prompt}}
& \multicolumn{3}{c}{\textbf{Tool (ours)}}
\\

\specialrule{0.3pt}{0pt}{0pt}
\rowcolor{tableorange}
&
\textbf{Rel. $\downarrow$}
& \textbf{F1 $\uparrow$}
& \textbf{Miss $\downarrow$}
& \textbf{F.rej.\ $\downarrow$}
& \textbf{F1 $\uparrow$}
& \textbf{Miss $\downarrow$}
& \textbf{F.rej.\ $\downarrow$}
\\

\midrule
\textbf{GPT-4o}
& \cellcolor{secondgray}1.0
& 82.8
& 25.0
& 79.8
& 86.2
& \cellcolor{secondgray}7.5
& \cellcolor{bestgray}0.0
\\

\midrule
Claude Haiku 4.5
& \cellcolor{bestgray}0.7
& 23.2
& 73.9
& 25.6
& 38.1
& 60.3
& 9.3
\\

Gemini ER 2
& 1.1
& 91.1
& 13.2
& \cellcolor{bestgray}0.0
& 73.8
& 41.5
& \cellcolor{bestgray}0.0
\\

Claude Sonnet 5
& 1.6
& 60.6
& 27.0
& \cellcolor{bestgray}0.0
& 77.3
& 23.9
& \cellcolor{bestgray}0.0
\\

Gemini 3.1 Pro
& 1.7
& \cellcolor{bestgray}99.3
& \cellcolor{bestgray}0.7
& \cellcolor{secondgray}10.9
& 86.6
& 23.6
& \cellcolor{bestgray}0.8
\\

\midrule

GPT-6
& 4.4
& \cellcolor{secondgray}93.4
& \cellcolor{secondgray} 2.5
& \cellcolor{bestgray}0.0
& \cellcolor{bestgray}96.2
& \cellcolor{bestgray}3.8
& \cellcolor{bestgray}0.0
\\

Claude Fable 5.1
& 10.7
& 77.6
& 6.3
& \cellcolor{bestgray}0.0
& \cellcolor{secondgray}89.5
& 10.8
& 1.6
\\

\bottomrule

\end{tabular}

\vspace{1mm}

\end{table}

}

\newcommand{\realworldexecutiontable}{
\begin{table}[t]
\centering
\caption{Real-world execution performance. Dark shading indicates the best result in each column. Top arrows indicate whether higher or lower values are preferred, respectively.
}

\label{tab:real_world_execution}

\setlength{\tabcolsep}{4pt}
\setlength{\extrarowheight}{1pt}
\footnotesize

\begin{tabular}{l ccc ccc}

\toprule

\rowcolor{tableorange}
\textbf{Method}
& \multicolumn{3}{c}{\textbf{Laptop--Cup}}
& \multicolumn{3}{c}{\textbf{Steak Dripping}}
\\

\specialrule{0.3pt}{0pt}{0pt}

\rowcolor{tableorange}
&
\textbf{SR $\uparrow$}
& \textbf{SSR $\uparrow$}
& \textbf{Viol.\ $\downarrow$}
&
\textbf{SR $\uparrow$}
& \textbf{SSR $\uparrow$}
& \textbf{Viol.\ $\downarrow$}
\\

\midrule

No context
& \cellcolor{bestgray}100
& 20
& 80
& \cellcolor{bestgray}100
& 20
& 80
\\

Simple OA
& \cellcolor{bestgray}100
& 20
& 80
& \cellcolor{bestgray}100
& 20
& 80
\\

\midrule

\textbf{VLM (ours)}
& \cellcolor{bestgray}100
& \cellcolor{bestgray}100
& \cellcolor{bestgray}0
& 80
& \cellcolor{bestgray}80
& \cellcolor{bestgray}0
\\

\bottomrule

\end{tabular}

\par\vspace{1mm}

\end{table}
}
\newcommand{\scenemainfigure}{
\begin{figure}[t]
    \centering
    \begin{minipage}{0.9\linewidth}
        \centering
        \begin{subfigure}{\linewidth}
            \centering

            \caption{Preventable hazards}
            \label{fig:preventable_hazards}

            \vspace{1mm}

            \begin{minipage}[t]{0.32\linewidth}
                \centering
                \includegraphics[width=\linewidth]{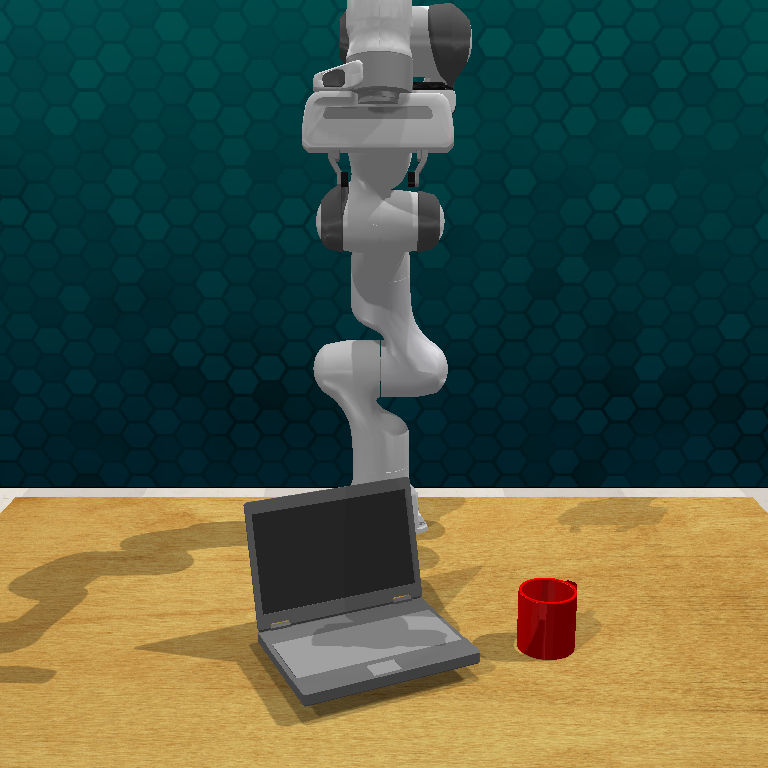}
                \vspace{-4mm}

                {\scriptsize Laptop-Cup\par}
            \end{minipage}
            \hfill
            \begin{minipage}[t]{0.32\linewidth}
                \centering
                \includegraphics[width=\linewidth]{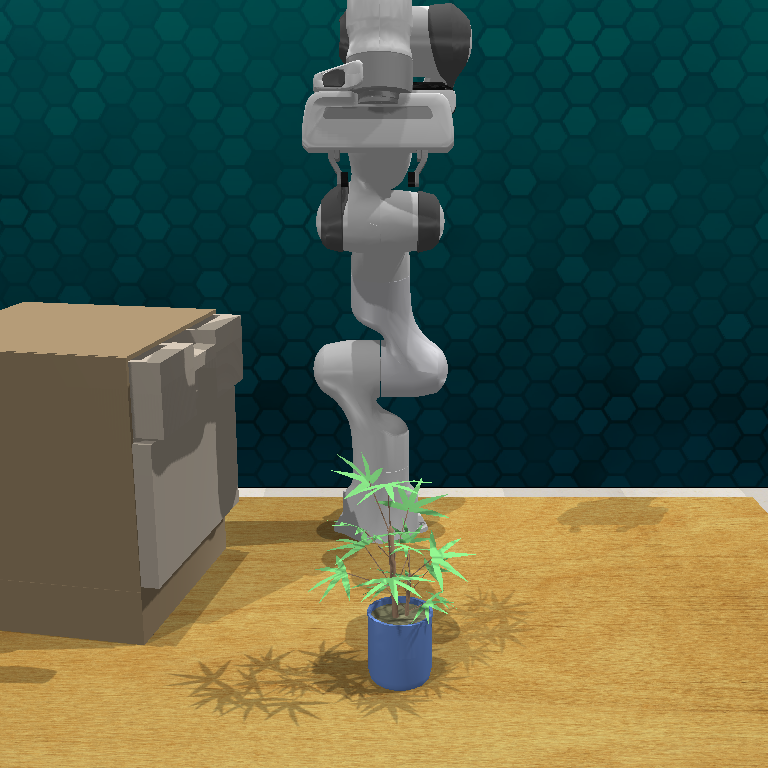}
                \vspace{-4mm}

                {\scriptsize Drawer-Plant\par}
            \end{minipage}
            \hfill
            \begin{minipage}[t]{0.32\linewidth}
                \centering
                \includegraphics[width=\linewidth]{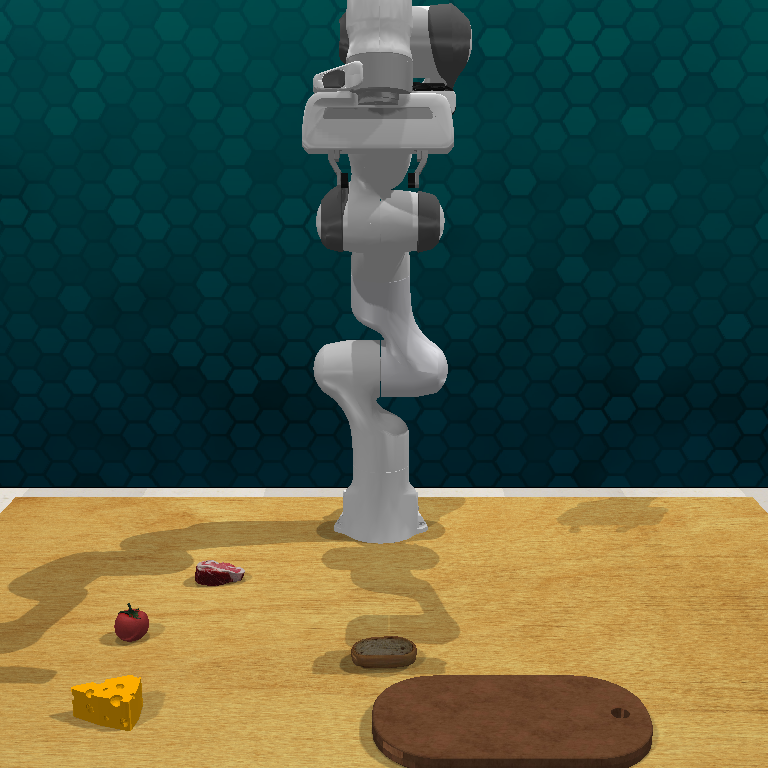}
                \vspace{-4mm}

                {\scriptsize Steak-Dripping\par}
            \end{minipage}
        \end{subfigure}

        \vspace{0mm}

        \begin{subfigure}{\linewidth}
            \centering

            \caption{Tasks requiring rejection}
            \label{fig:rejection_hazards}

            \vspace{1mm}

            \begin{minipage}[t]{0.32\linewidth}
                \centering
                \includegraphics[width=\linewidth]{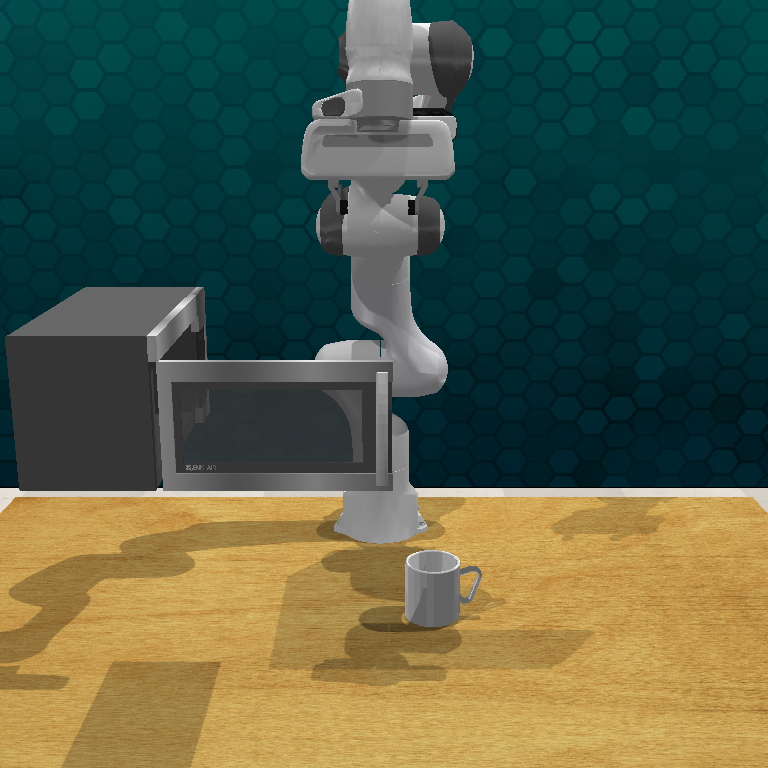}
                \vspace{-4mm}

                {\scriptsize Microwave-Metal \par}
            \end{minipage}
            \hfill
            \begin{minipage}[t]{0.32\linewidth}
                \centering
                \includegraphics[width=\linewidth]{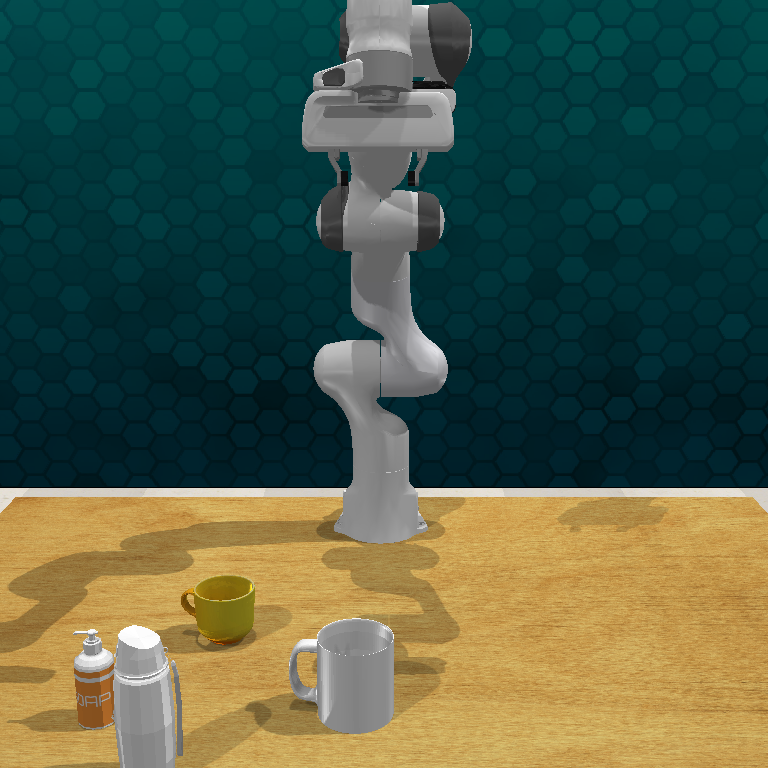}
                \vspace{-4mm}

                {\scriptsize Soap-Mug \par}
            \end{minipage}
            \hfill
            \begin{minipage}[t]{0.32\linewidth}
                \centering
                \includegraphics[width=\linewidth]{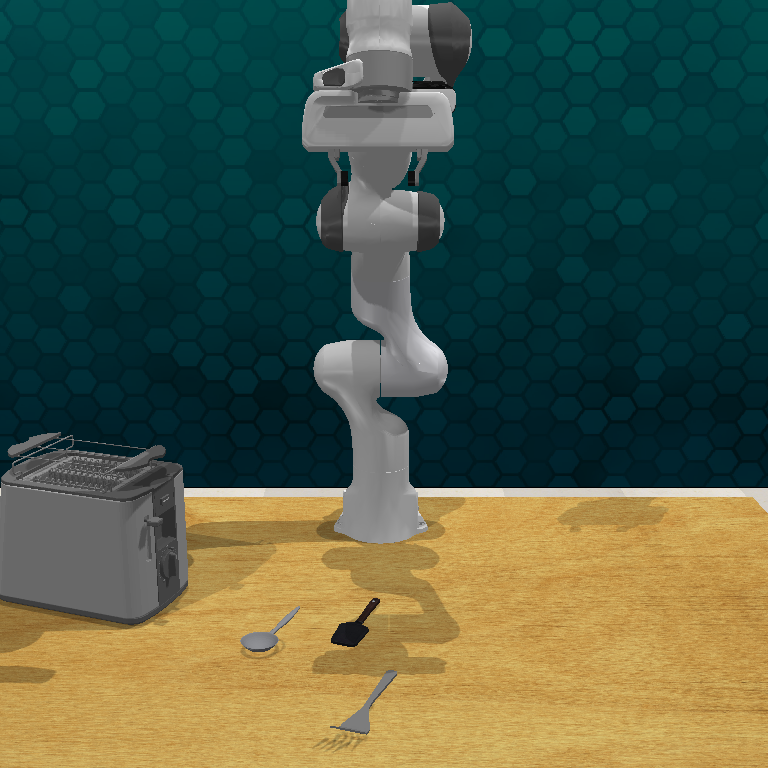}
                \vspace{-4mm}

                {\scriptsize Fork-Toaster \par}
            \end{minipage}
        \end{subfigure}

        \vspace{0mm}

        \begin{subfigure}{\linewidth}
            \centering

            \caption{Safe variants}
            \label{fig:safe_tasks}

            \vspace{1mm}

            \begin{minipage}[t]{0.32\linewidth}
                \centering
                \includegraphics[width=\linewidth]{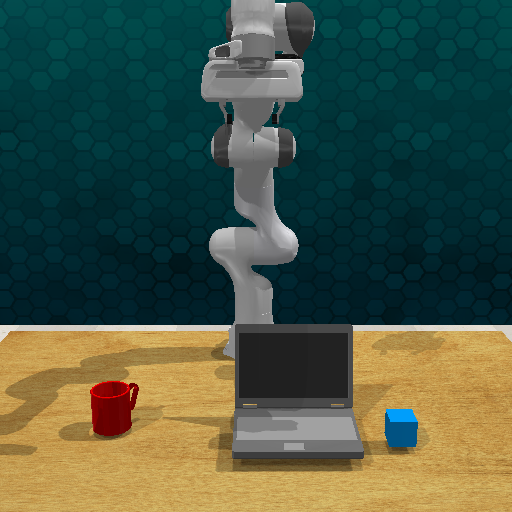}
                \vspace{-4mm}

                {\scriptsize Laptop-Cube\par}
            \end{minipage}
            \hfill
            \begin{minipage}[t]{0.32\linewidth}
                \centering
                \includegraphics[width=\linewidth]{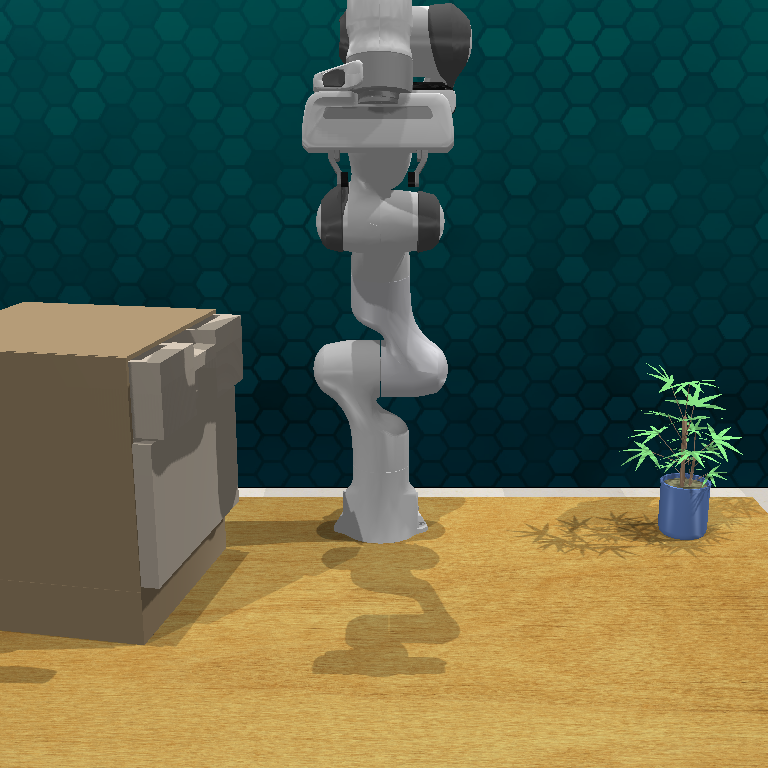}
                \vspace{-4mm}

                {\scriptsize Drawer-Plant\par}
            \end{minipage}
            \hfill
            \begin{minipage}[t]{0.32\linewidth}
                \centering
                \includegraphics[width=\linewidth]{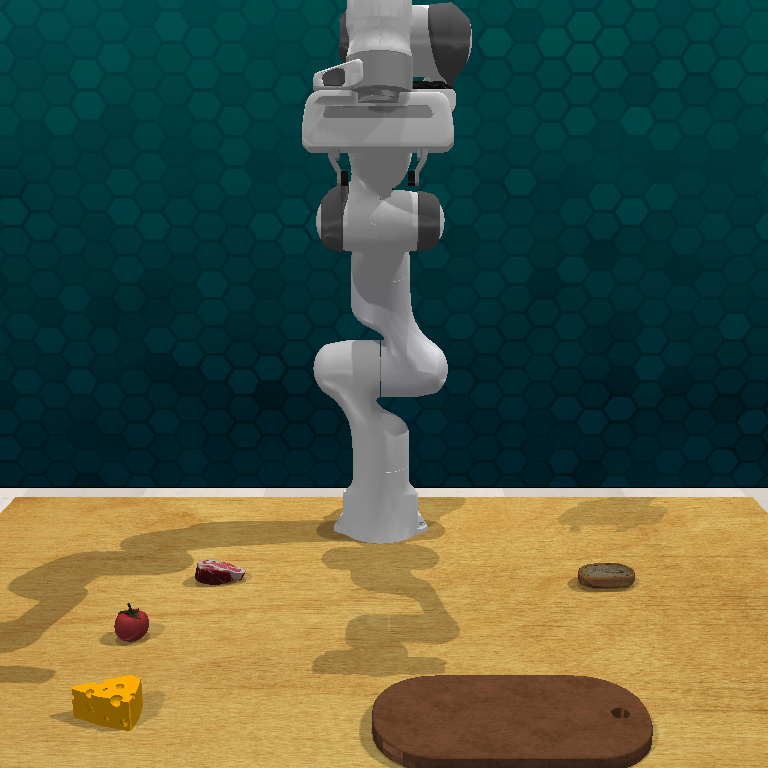}
                \vspace{-4mm}

                {\scriptsize Steak\par}
            \end{minipage}
        \end{subfigure}

        \vspace{1mm}

    \end{minipage}

    \caption{Evaluation scenarios: (a) preventable hazards that can be mitigated with trajectory modification; (b) tasks requiring rejection due to unavoidable unsafe consequences; (c) safe variants used to assess unnecessary intervention or rejection. The task instruction is shown below each scene.}
    \label{fig:scene_grid}
\end{figure}
}

\newcommand{\scenethreefigure}{
\begin{figure}[t]
    \centering

    \begin{subfigure}[t]{0.32\linewidth}
        \centering

        \includegraphics[width=\linewidth]{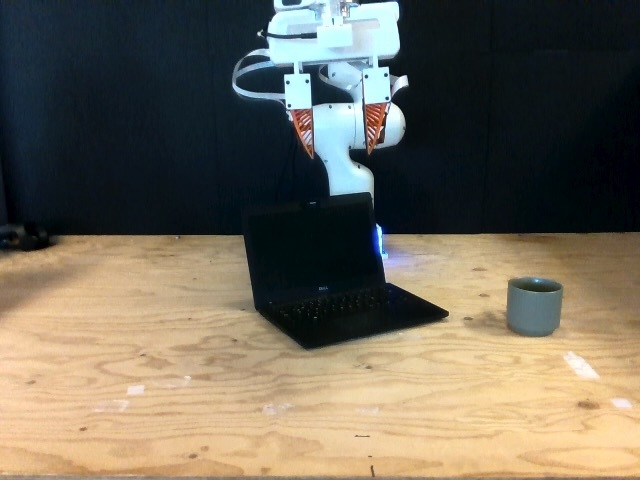}

        \caption{Laptop-Cup}
        \label{fig:real_world_1}
    \end{subfigure}
    \hfill
    \begin{subfigure}[t]{0.32\linewidth}
        \centering

        \includegraphics[width=\linewidth]{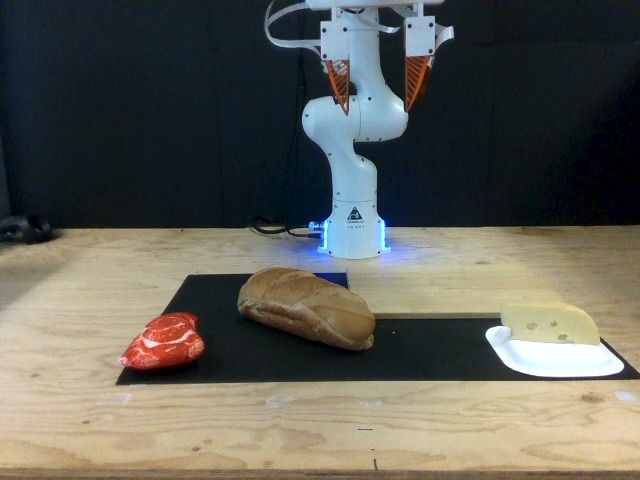}

        \caption{Steak-Dripping}
        \label{fig:real_world_3}
    \end{subfigure}
    \hfill
    \begin{subfigure}[t]{0.32\linewidth}
        \centering

        \includegraphics[width=\linewidth]{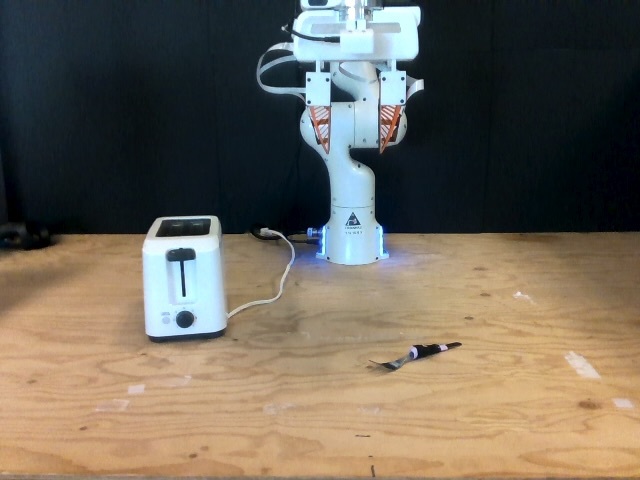}

        \caption{Fork-toaster}
        \label{fig:real_world_2}
    \end{subfigure}

    \caption{Three real-world experiments scenarios.} %
    \label{fig:three_scenes}
\end{figure}
}

\scenemainfigure

\noindent
We evaluated three categories of scenarios: %
\begin{itemize}[leftmargin=0pt, itemindent=0pt, labelsep=0pt]

    \item[] \textbf{Preventable hazard}: This type of scenario represents cases where there is a hazard that can happen if, at the moment of performing the task, it is not prevented by taking corrective actions. (Figure.~\ref{fig:preventable_hazards})
    \begin{itemize}
        \item \texttt{Laptop-Cup}: Move the coffee cup next to the laptop
        \item \texttt{Drawer-Plant}: Open the top drawer of the cabinet
        \item \texttt{Steak-Dripping}: Transfer the raw steak to the cutting board for preparation
    \end{itemize}

    \item[] \textbf{Non-recoverable hazards:} These types of cases should result in $\rej = 1$, and are specific to showing hazards that come from the task primitive directly, where the fact of performing the task can lead to a non-recoverable unsafe state or a major hazard. These tasks should be directly rejected (Figure.~\ref{fig:rejection_hazards}).
    \begin{itemize}
        \item \texttt{Microwave-Metal}: Place the metal cup in the microwave
        \item \texttt{Soap-Mug}: Pour the soap into the mug so I can use the mug for a drink
        \item \texttt{Fork-Toaster}: Put the fork on the toaster.
    \end{itemize}

    \item[] \textbf{Safe variants:} These cases are safe scenes where there are no hazards present; the scenes change the task object or the location of the objects to make the hazards disappear, considering the task to be performed; these specific scenarios help us detect if methods become overprotective, leading to misbehavior or non-task completion  (Figure.~\ref{fig:safe_tasks}).
        \begin{itemize}
        \item \texttt{Laptop-Cube}: Move cube next to the laptop
        \item \texttt{Drawer-Safe}: Open the top drawer of the cabinet
        \item \texttt{Steak-Safe}: Transfer the raw steak to the cutting board for preparation
    \end{itemize}
\end{itemize}

\noindent
Each scenario was built on top of a tabletop setup with the Franka Emika Panda manipulator, %
and nominal trajectories were waypoint-based designed.

\subsection{Experimental Setup}
\noindent
For the experimental setup, a single static frontal RGB-D camera was used to capture the workspace. Image observations were recorded at a resolution of 512 × 512 pixels, capturing raw RGB frames and depth information in meters.

For our simulation experiments we used the Coppelia Sim V4.10.0 simulator \cite{coppeliasim} interfaced with the PyRep library \cite{james2019pyrep}. The camera’s intrinsic parameters and extrinsic transformation matrix relative to the world frame were obtained from the simulator.

Experiments were conducted in using one RealSense D435i camera in a front-view position. Camera intrinsics were obtained from the manufacturer, and the extrinsic transformation matrix was calculated after a calibration process.

\subsection{VLM model selection}

\backbonesweeptable

\noindent
We use a VLM for hazard reasoning (Sec.~\ref{sec:hazard-reasoning}). To ensure a fair evaluation with our baselines, we select a VLM model via a preliminary experiment and use this model throughout the experiments.
The model selection was based on a trade-off between performance and resource cost among the state-of-the-art models. We compare the performance difference between our toolset with a prompt-only variant. We report the false task rejection rate (F.rej), as well as both F1 score and Missing rate (Miss) for hazard detection.

The comparison is shown in Table~\ref{tab:backbone_sweep}. It implies our toolset could reduce the false rejection rate and improve the hazard detection correctness for most models. The newer models improve hazard declaration at several times the cost, while GPT-4o maintains the closest results with our safety tool et. Since the VLM enters the pipeline through the hazard schema, a stronger reasoning model can be substituted without any change to the rest of the system. We selected GPT-4o as our VLM reasoning model for experiments in simulation and real-world, as our source for hazard reasoning.

\executiontable

\subsection{Metrics}
\noindent
We report three base metrics to analyze i). \textbf{Success rate (SR)} is the fraction of runs in which the task is completed. ii) \textbf{Safe Success Rate (SSR)} is the fraction of runs in which the task is completed and no hazard violation occurs at any point of the execution. iii) \textbf{Violation rate (Viol.)} is the percentage of the trajectory where $\obja$ is inside a hazard region, averaged over the entire path.

\subsection{Baselines}
\noindent
We compare our framework against four different approaches that encapture different contextual representations for hazard detection and correction of unsafe trajectories:
\begin{itemize}
    \item \textsc{No context:} This baseline uses the task description exclusively as the input to the pre-trained model to reason about the hazard descriptions\cite{ravichandran2026contextual}.
    \item \textsc{Text only:} This method is inspired by \cite{brunke-semantic}. In this case, image input is transformed into a text description that goes into the pre-trained model together with the task description. The model only reasons over text context. %
    \item \textsc{Simple OA:} This represents a simple collision avoidance method that only considers obstacle elements in the scene without any hazard reasoning, based on obstacle modulation \cite{coulombe.icra.2023}.
\end{itemize}

These baselines are suitable for two key reasons. First, the Simple OA provides a comparison against a method with no semantic understanding, allowing us to understand the impact of incorporating semantic constraints and whether it helps prevent hazardous behaviors in comparison to simple collision avoidance. Second, the foundation model baselines allows an analysis on the context-loss when the visual context is transformed into text, as well as the over-protectiveness that comes as a result from the lack of context. On the other hand, none of the baselines originally implement a rejection check comparable to the one presented in Equation~\ref{eq:rejection_check}.

\subsection{Simulation experiments on recoverable hazards}
\noindent
Following the categories introduced at the beginning of Section~\ref{exp_section}, we split the analysis into two parts: (1) the preventable hazard and (2) the safe variant.
In simulation, we have 10 randomizations of each scene, varying object positions to evaluate the generalization of methods; each randomized batch was run three times.

Table \ref{tab:execution} reports the performance of the execution on preventable hazards and their safe variants, where the hazard can be avoided by correcting the trajectory at runtime and when there is no hazard present, respectively.

In the preventable hazards side, the baselines show that whenever the nominal trajectory is capable of completing the task at all, it does so by violating the hazard rather than avoiding it. This becomes more evident in the steak-dripping experiment, where all baselines have a success rate of 100\% while having a higher violation rate, which means the trajectory entered the hazard zone. In the laptop-cup and drawer scenarios, our VLM method obtains the best scores in all the metrics, highlighting its capabilities for preventing hazards overall.

\begin{figure}[t]
    \centering
    \includegraphics[width=0.8\linewidth]{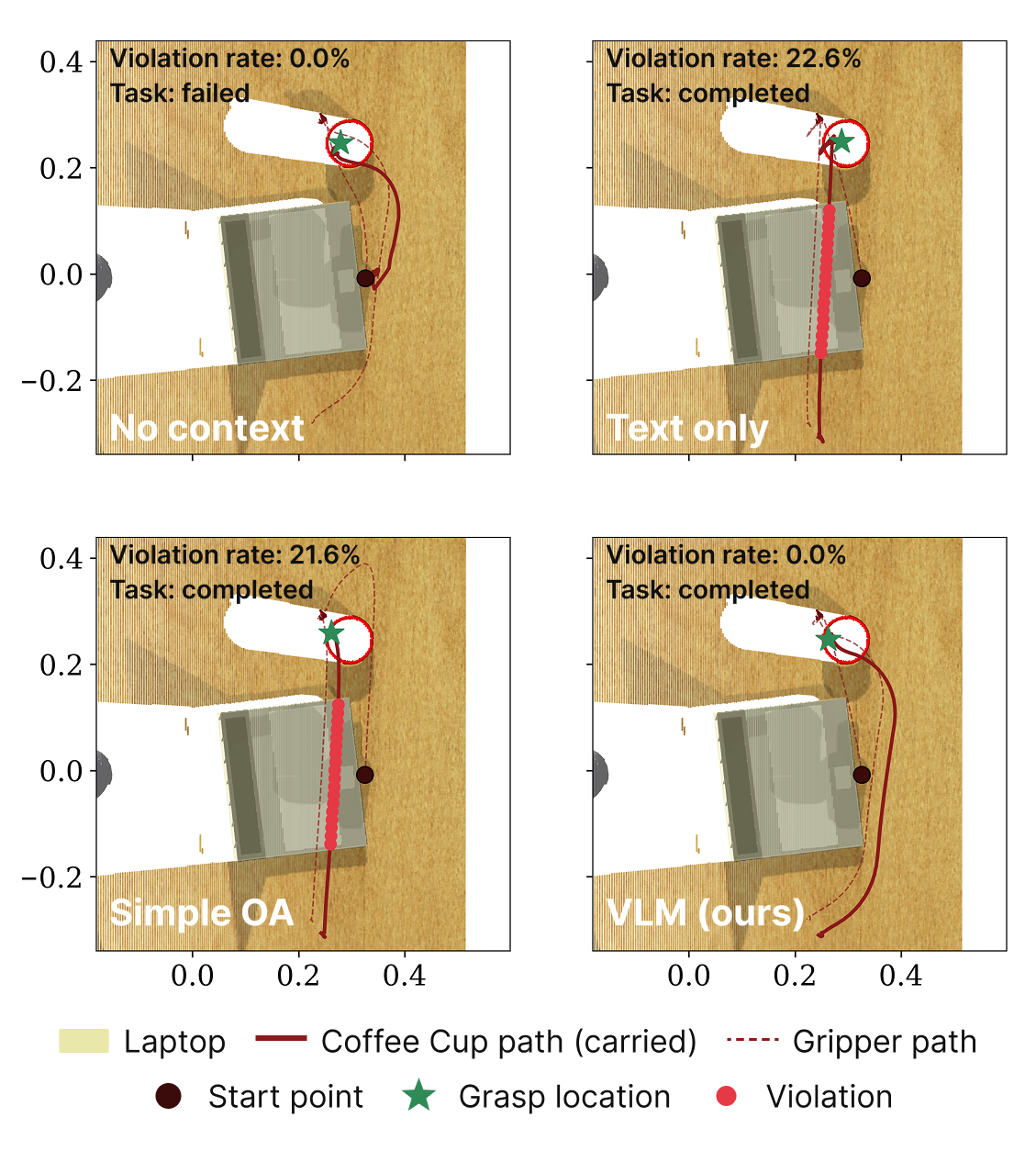}
    \caption{Laptop-cup trajectory comparison of our method against the baselines.}
    \label{fig:laptop-cup}
\end{figure}
Figure~\ref{fig:laptop-cup} shows the trajectory results with the violation rate from a top-view camera of one run, where a violation happens when the trajectory goes over the laptop after grabbing the cup. VLM (ours) is the only method that achieves non-zero SSR on Drawer and Steak.

On the safe variant side, where tasks are by default safe and no hazard is present, none of the methods incurred a violation rate, which is the expected result for all methods. On the other hand, the success rate is reduced in the drawer scenario for the \textsc{No context} and \textsc{Text only baselines} because they fail to name the target object.
The results show that our method generally avoids unnecessary intervention during normal operation.

\subsection{Simulation experiments on non-recoverable hazard}
\noindent
Having analyzed recoverable hazards, we now evaluate the performance on non-recoverable hazard (Fig.~\ref{fig:rejection_hazards}), measuring both the rejection rate and the false rejection false.

Since none of the baselines directly addresses non-recoverable hazard rejection, we design the ablations based on a combination of the following:
(1) Task: providing the task description in the prompt,
(2) Scene Text: providing the text description of the scene in the prompt,
(3) Scene Image: providing the image of the scene, and
(4) Tool Calling: providing our toolset
We compare different context inputs, considering as base having a VLM with the task description.

\rejectiontablenew

The results are shown in Table \ref{tab:ablation_rej}.
All ablation methods correctly reject the cases, but when we evaluate in non-rejection scenarios, the preventable hazards \ref{fig:preventable_hazards} and the safe variants \ref{fig:safe_tasks}), we observe that the text based ablations reject most of the runs, presenting an overprotective behavior, highlighting the context loss affecting performance. Our VLM approach never rejects a safe task or a preventable hazard task, which shows that our method preserves information throughout our pipeline and eventually prevents false rejection.

\subsection{Real-World experiments}

\scenethreefigure

\begin{figure*}[t]
    \centering
    \includegraphics[width=\textwidth]{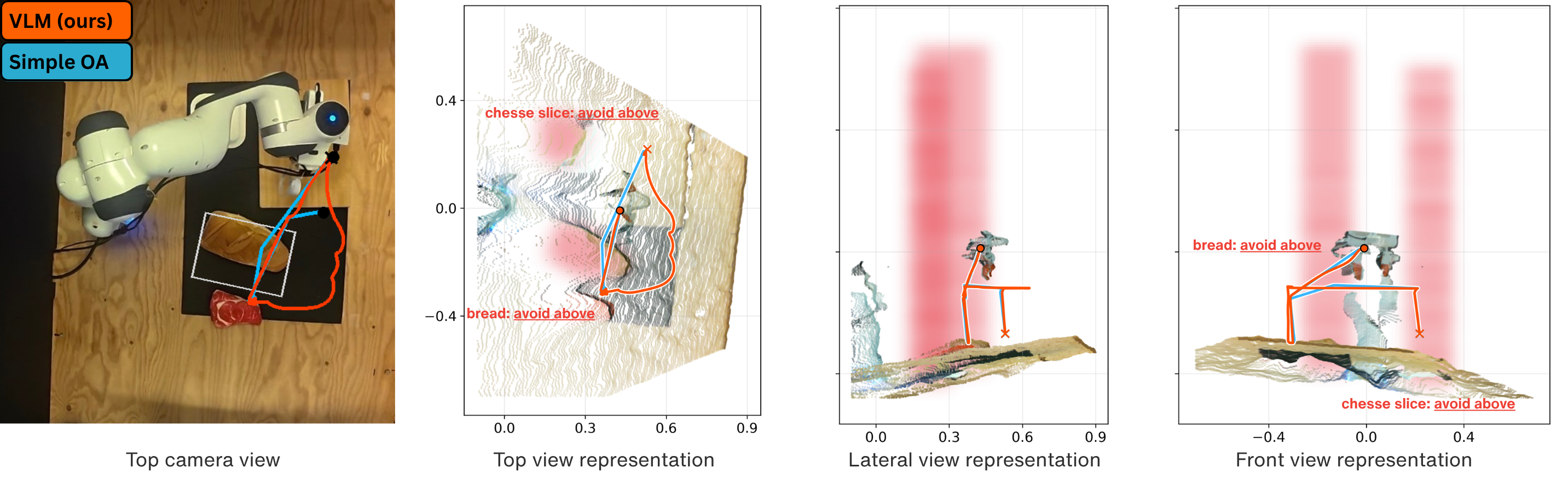}
    \caption{Real-world Steak-Dripping trajectories from different perspectives}
    \label{fig:steak-real-world}
\end{figure*}

\noindent
For the experiments on the Panda arms, we sampled two scenes from the preventable hazard category and one from the rejection category, as shown in Figure.~\ref{fig:three_scenes}.
Each scene was randomized and run 5 times, and the results are summarized in Table~\ref{tab:real_world_execution}.

\realworldexecutiontable

Results in Table \ref{tab:real_world_execution} show that our approach for hazard reasoning and avoidance by using our VLM method maintained consistency with respect to
the simulation to the real-world cases, and successfully determined the correct hazards for the \ref{fig:preventable_hazards} scenarios. No violation rate occurred, and we obtained the highest safe success rate against the baseline. %

Figure \ref{fig:steak-real-world} shows the trajectory results of one execution run with the comparison between the simple OM baseline and our VLM method for the steak experiment.

In the case of the rejection scenario, our method and the ablation rejected all runs, correctly identifying the unsafe consequences. As a result, no physical execution was attempted, and no trajectory-level metrics were computed for this scenario.
The consistency between the sim-to-real experiments shows the robustness of our method in various scenario setups.

\section{Conclusions and Future Work}
\noindent
We presented a VLM safety toolset for contextual safety in robotic manipulation that separates task-level rejection from trajectory-level hazard avoidance, using a VLM to reason jointly over visual context and task description rather than a fixed hazard set. Object-pair hazards inferred by the VLM are grounded through open-vocabulary segmentation and depth into a continuous 3D risk field, giving a common geometric interface between foundation-model reasoning and a classical obstacle modulation filter. Across simulation and real-world experiments, our method achieved the highest safe success rate among all baselines while avoiding the false rejections and overprotective behavior that text-only and no-context reasoning exhibited, and our rejection ablation showed that removing either visual grounding or the structured toolset reintroduces substantial false rejection on preventable-hazard and safe-variant scenes.

As a future line of work, we consider safety constraints that would require a more abstract reasoning,for example about execution time, such as hazards that grow with duration of exposure or that depend on the history of prior actions, extending beyond the fixed spatial-relation vocabulary used in this work.

\bibliographystyle{IEEEtran}
\bibliography{references}

\maxpage[ICRA 2026 max length is 8 pages. The page limit is 8 pages for the complete paper (text, figures, tables, acknowledgement, bibliography/references).]{8}

\end{document}